# From Density to Geometry: YOLOv8 Instance Segmentation for Reverse Engineering of Optimized Structures

Thomas Rochefort-Beaudoin[a], Aurelian Vadean[a], Sofiane Achiche[a], Niels Aage[b]

[a]Department of Mechanical Engineering, Polytechnique Montréal, 2500 Chemin de Polytechnique, Montréal, QC, Canada
[b]Department of Civil and Mechanical Engineering, Solid Mechanics, Technical University of Denmark. Koppels Allé, B.404, 2800 Kgs. Lyngby, Denmark

## Abstract

This paper introduces YOLOv8-TO, a novel approach for reverse engineering of topology-optimized structures into interpretable geometric parameters using the YOLOv8 instance segmentation model. Density-based topology optimization methods require post-processing to convert the optimal density distribution into a parametric representation for design exploration and integration with CAD tools. Traditional methods such as skeletonization struggle with complex geometries and require manual intervention. YOLOv8-TO addresses these challenges by training a custom YOLOv8 model to automatically detect and reconstruct structural components from binary density distributions. The model is trained on a diverse dataset of both optimized and random structures generated using the Moving Morphable Components method. A custom reconstruction loss function based on the dice coefficient of the predicted geometry is used to train the new regression head of the model via self-supervised learning. The method is evaluated on test sets generated from different topology optimization methods, including out-of-distribution samples, and compared against a skeletonization approach. Results show that YOLOv8-TO significantly outperforms skeletonization in reconstructing visually and structurally similar designs. The method showcases an average improvement of 13.84% in the Dice coefficient, with peak enhancements reaching 20.78%. The method demonstrates good generalization to complex geometries and fast inference times, making it suitable for integration into design workflows using regular workstations. Limitations include the sensitivity to non-max suppression thresholds. YOLOv8-TO represents a significant advancement in topology optimization post-processing, enabling efficient and accurate reverse engineering of optimized structures for design exploration and manufacturing.

## 1. Introduction

Topology Optimization (TO) is used by engineers to improve the design process, by leveraging numerical methods to seek the optimal material distribution within a given domain, while maximizing key properties such as stiffness or heat transfer in the resulting structure.

Density-based TO [1], such as the Solid Isotropic Material with Penalization (SIMP) method [2], employs a discretization approach where the structural domain, outlining the structure's maximum bounds, is divided into a grid of pixels (or voxels for 3D). Each pixel assumes a continuous density value, ranging from 0 (void) to 1 (presence of material). The SIMP method has become the de-facto framework in the industry and academia due to its simplicity of implementation [3], scalability for large-scale problems [4], [5], support for a wide range of manufacturing constraints [6], and optimization objectives [7].

The density distribution matrix encodes the spatial distribution of material inside the given domain. Therefore, the optimal structure's geometry is implicitly represented by the assembly of pixel values close to 1, representing the presence of material. TO results often contain several millions of density variables and lead to highly complex material layouts with a large number of structural features. This means that human interpretation and feature extraction is extremely time-consuming, if possible at all. Therefore, automatic post-processing steps are required to interface the resulting structure with other steps in the mechanical design process [8].

The literature on the geometric post-processing of topology-optimized structures can be separated into three distinct downstream applications: prototyping, design validation, and design exploration [9]. While the first two applications are focused on generating a high-quality surface mesh for 3D printing or finite-element analysis (FEA), design exploration applications require a parametric output that can be manipulated by the engineer to iterate toward the optimal design. Ideally, the density field is converted to a set of interpretable geometrical parameters which is compatible with CAD software to streamline the design process and facilitate design iterations.

To address this, feature-mapping methods [10] for TO were developed to directly optimize geometric variables defining high-level geometric components which are then mapped onto the density field via some projection methods. The Method of Moving Morphable Components (MMC) [11] uses hyperelliptic functions defined by center coordinates, length, orientation, and thicknesses to build a density distribution out of an assembly of the high-level geometric primitives which are then projected onto a discretized domain. These methods allow direct interpretability of the design into simple geometric variables but currently show worse convergence rates [12] than the well-established density-based methods due to being more prone to getting stuck in poor local minima [10].

Ideally, the geometric post-processing for design exploration should be done independently of the optimization process used and can be done as a reverse-engineering task where a structure is reconstructed from a set of high-level geometries predefined by the engineer. By performing the geometric post-processing as a separate step, well-established and efficient density-based methods can be used for optimization, ensuring faster convergence and better performance. The resulting optimized density distribution can then be reverse engineered into a set of high-level geometric primitives, which are more intuitive and easier to manipulate by the engineer. This two-step approach combines the strengths of both density-based methods and feature-mapping methods, allowing for efficient optimization and interpretable geometric representation of the optimized structure, ultimately streamlining the design process, and facilitating design iterations.

The field of reverse-engineering of 3D point clouds into Constructive Solid Geometry (CSG) trees is well-documented in the literature [13], [14], [15]. However, the specific application of these methods to TO outputs within a design exploration framework has been limited to skeletonization techniques. These techniques essentially involve finding a one-dimensional representation of the optimal density distribution. Despite their relevance, these methods underperform for structures that do not resemble trusses [16], [17], [18]. Even advanced methods using curved skeleton members, which are better suited for TO outputs with complex shapes, struggle with accurately locating the intersections of skeleton branches [19]. Furthermore, the most effective skeletonization methods often require manual inputs and are reliant on heuristics specific to each case [20]. Even so, skeletonization-based methods used to convert density-based outputs to MMC design variables have demonstrated the added potential of accelerating convergence of conventional MMC optimization by offering the possibility to set the initial solution from an interpreted intermediate SIMP topology, resulting in the reduction of computation time from 16% to 23% [21].

The development of a reverse engineering method for TO can be compared to creating a geometrical compression algorithm for encoding complex 2D or 3D density distributions into a set of one-dimensional geometric parameters as illustrated in Figure 1. The key challenge is to ensure that this 1D representation allows for almost lossless decoding back to the original 2D or 3D form, hence preserving essential features of the optimized design. The effectiveness of this process should be evaluated both at the image level (per-pixel accuracy) and the mechanical level (structural response under load).

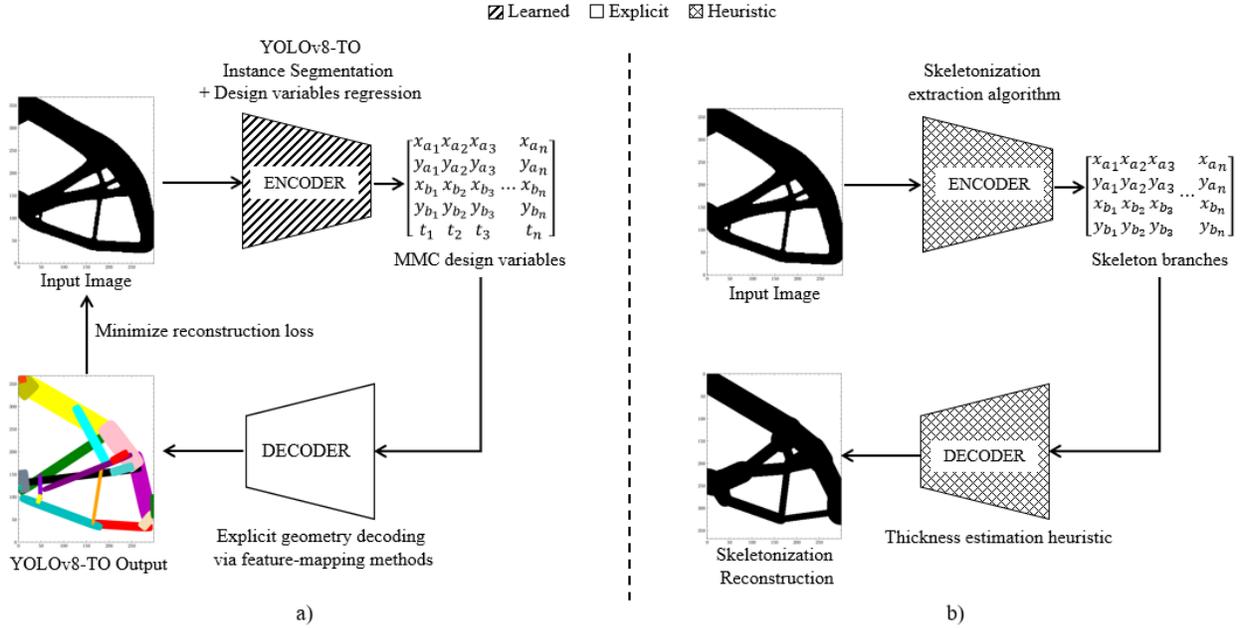

Figure 1: Compression algorithm analogy. a) YOLOv8-TO as a learnable geometrical encoder with an explicit geometry decoding defined by the MMC hyperelliptic functions. b) Skeleton extraction heuristic as a geometrical encoder with a heuristic-based thickness estimation decoder.

This article proposes a novel approach for reverse-engineering TO results by framing the problem as an instance segmentation task. Instance segmentation is a computer vision technique that involves detecting and delineating distinct objects of interest within an image, while simultaneously classifying each object. By leveraging the YOLOv8 algorithm, our method integrates image segmentation with design variable regression, focusing on accurately identifying and reconstructing component shapes within monochrome images. This approach represents an efficient geometric encoding method with an explicit decoding algorithm, offering a more straightforward and robust alternative to heuristics-based skeletonization approaches. The novelty of this work lies in the tailored application of state-of-the-art instance segmentation techniques to the field of TO, enabling the automatic extraction of interpretable geometric parameters from TO results.

## 2. Method

The task of identifying critical structural members in a 2D density distribution can also be understood as an instance segmentation task with the specific requirement to output, for each detected instance, geometric variables defining its position and shape. Figure 2 illustrates the segmentation process that enables the reconstruction of each structural sub-structure in a TO output.

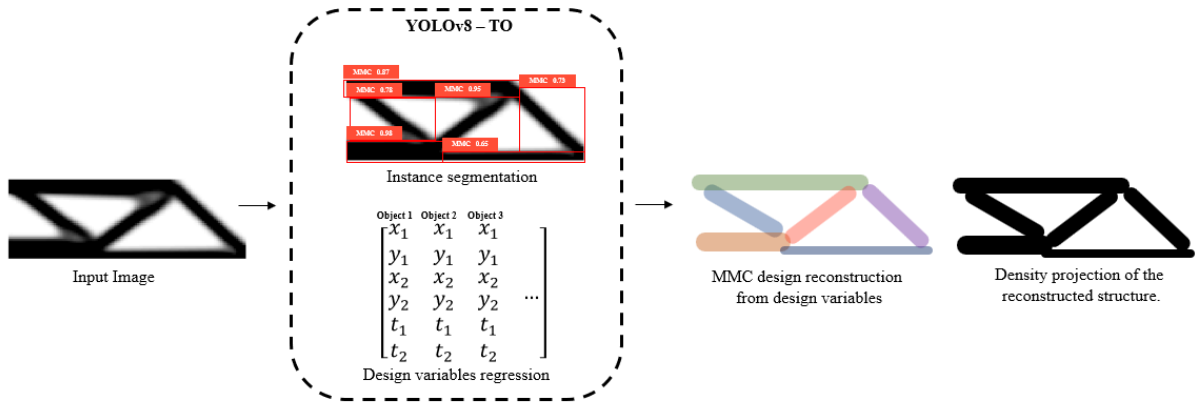

Figure 2: Reverse-engineering of an optimal density distribution as an instance segmentation approach.

The presented instance segmentation approach is based on YOLOv8 [22], a cutting-edge computer vision model focused on having low inference latency and presenting an open-source and well-maintained codebase. In addition to the object classification, bounding box prediction, and pixel segmentation heads, an additional regression head was added to the model for the task of design variables regression. The custom model is detailed in section 2.2.

## 2.1 Dataset

The model was trained and tested on a diverse set of datasets to evaluate its performance and generalization capabilities. The datasets include optimal topologies obtained from the MMC and SIMP methods, random assemblies, and out-of-distribution (OOD) samples from the literature. The MMC dataset was split into training, validation, and testing subsets, while the random assembly dataset was used for training only. The SIMP, low volume fraction SIMP (SIMP5%), and OOD datasets were used for testing.

### 2.1.1 MMC dataset

The MMC method was used as the basis of the training dataset because its output can be directly utilized as a segmentation label for the model training. The MMC implementation from [23] was used and a hybrid optimizer combining both the Method of Moving Asymptotes (MMA) and its globally convergent version (GCMMA) was chosen because of its better convergence behavior for similar problems in 2D [24]. To generate many samples, random boundary conditions, inspired by [25], were sampled from predefined distributions as presented in Table 1 and Figure 3. The black-and-white density projections obtained via a Heaviside projection were used as input for the instance segmentation model as illustrated in Figure 4. The dataset was split into training (80%), validation (10%) and testing (10%) subsets to evaluate the model's performance on optimal topologies.

Table 1: Parameters defining the boundary conditions distribution.

| Parameter | Name | Distribution |
|---|---|---|
| $h$ | Height | [1.0, 2.0] |
| $w$ | Width | [1.0, 2.0] |
| $L_s$ | Support length | 50% to 75% |
| $P_s$ | Support position | 0 to (100% of $L_s$) |
| $P_L$ | Load position | 0% to 100% of boundary opposite from support |
| $\theta_L$ | Load orientation | [0°,360°] * |

*The selected angle is filtered to ensure there is at least 45 degrees of difference with the support normal

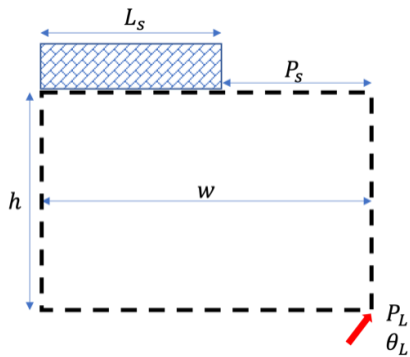

Figure 3: Illustration of the boundary conditions distribution

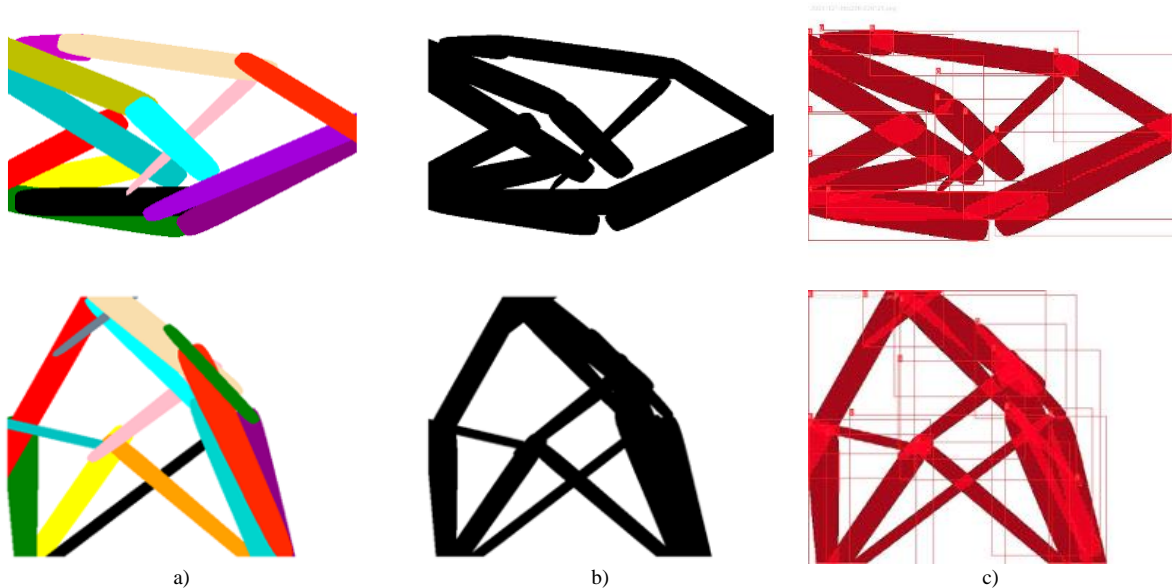

a)           b)           c)

Figure 4: MMC dataset training sample. a) Output from MMC optimization. b) Black and white input for the instance segmentation model. c) Instance segmentation labels.

### 2.1.2 Random assembly dataset

The generation of training data for machine learning models using iterative gradient-based TO methods is computationally expensive. This computational burden has constrained much of the existing research on machine learning applications in TO to rely on small, homogeneous datasets with low-resolution samples for training and testing [26], [25]. YOLOv8-TO's ability to learn from non-optimized structures is a key property as it can be trained on simple, binary images of structural components. Randomly sampling the space of design variables allows for the cost-effective generation of assemblies composed of randomly distributed components. The geometry function defining the components implicitly defines the segmentation labels required by the detection head of YOLOv8 while the random design variables are used as labels by the regression head. Samples of the random assembly dataset are shown in Figure 5.

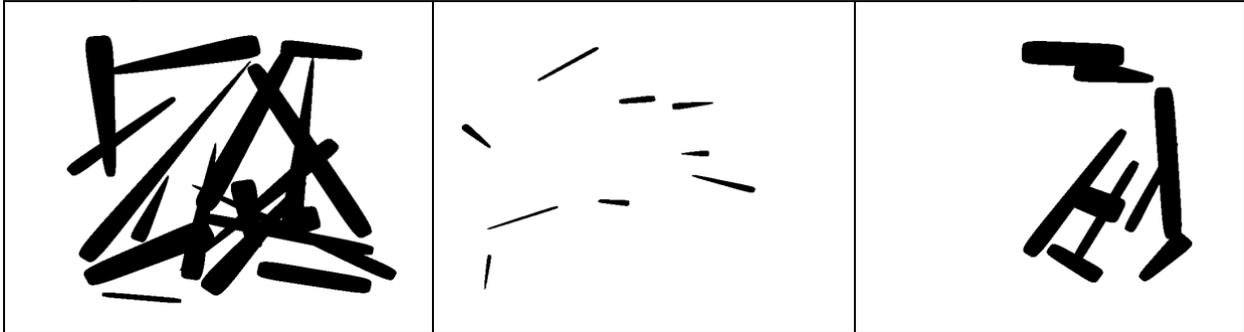

Figure 5: Random MMC training samples.

### 2.1.3 SIMP dataset

Testing the model on a dataset of optimal structure generated from a different TO method allows to test the capability of the method as a general post-processing tool, independent of the optimization technique used. A dataset of 2000 TO structures was generated using the boundary conditions sampled from the same distribution as the MMC dataset. Samples are presented in Figure 6. A mesh resolution of 150 elements per unit of height/width and the Python implementation from the 88-line MATLAB code for TO were used [27].

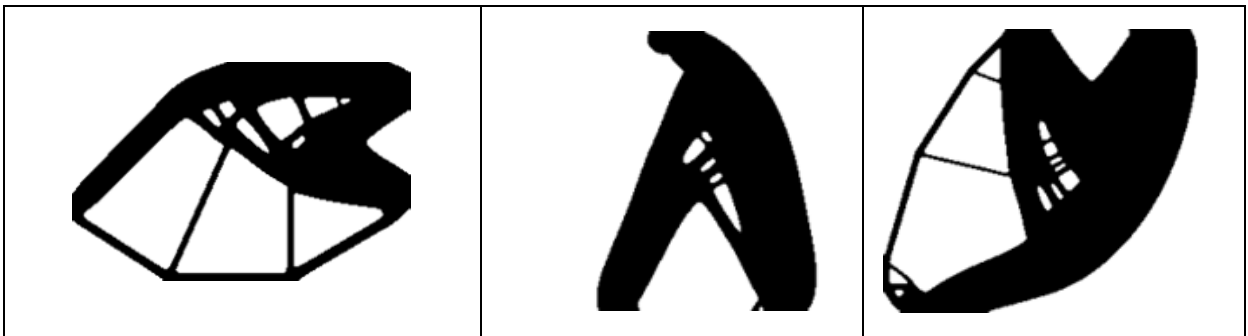

Figure 6: SIMP test dataset samples.

### 2.1.4 Low volume fraction SIMP dataset (SIMP$_{5\%}$)

To compare the performance of YOLOv8-TO against a skeletonization approach, 2000 random SIMP samples with the same boundary conditions were generated with a low volume fraction of 5%. This generated thin structures that better fit the "truss-like" property which is better suited for skeletonization approaches. To obtain a binary density distribution, as shown in Figure 7, a threshold value of 0.1 was applied to the generated continuous density values. All density values above 0.1 were set to 1.0, while values equal to or below 0.1 were set to 0.0. This dataset will be referred to as 'SIMP$_{5\%}$'.

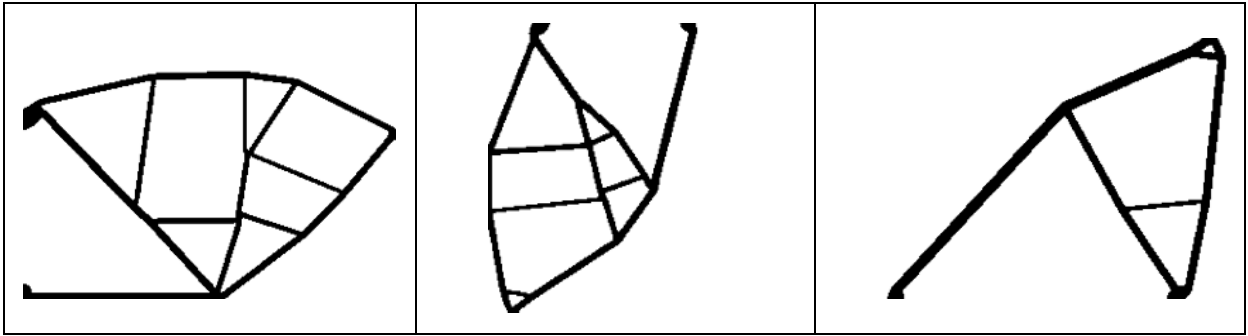

Figure 7: Low volume fraction SIMP samples with volume constraint of 5%.

### 2.1.5 Out-of-distribution (OOD) dataset

To test the generalization capability of the proposed approach, 4 images of TO structures taken from the literature were selected to test against some complex structure types that were not encountered in the training phase. Figure 8a) and b) depict a 2D femur structure that has been optimized using a density-based SIMP method, subject to total volume constraints and local volume constraints respectively [28]. These two images enable the evaluation of the model's ability to interpret curved and porous sub-structures. Figure 8c) represents the classic 2D cantilever beam problem partially optimized via the level set method [29]. The image was generated using the 88-line level set educational code [30] and was limited to 20 optimization iterations to obtain thick and curved structural members which are different from the straight MMC components used in training. Figure 8d) corresponds to a compliant mechanism force inverter obtained with the 88-lines SIMP code [27].

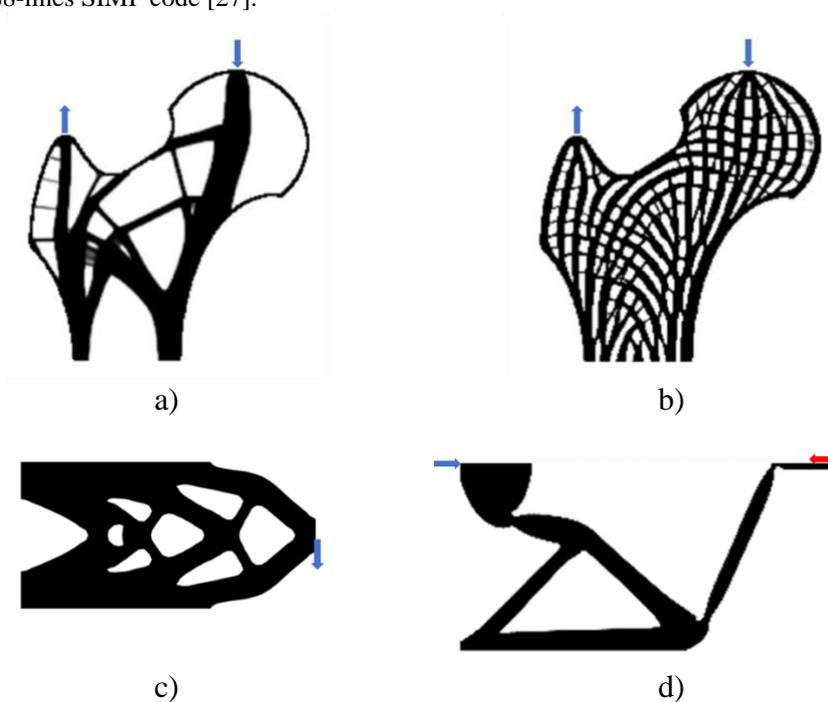

Figure 8: Out-of-distribution test samples. a): 2D femur with total volume constraint. b) 2D femur with local volume constraints, reproduced with permission from [28]. c) Cantilever beam partially optimized with the level set method, obtained with [30]. d) Compliant mechanism force inverter, obtained with [27]. The red arrow shows the optimization objective to maximize.

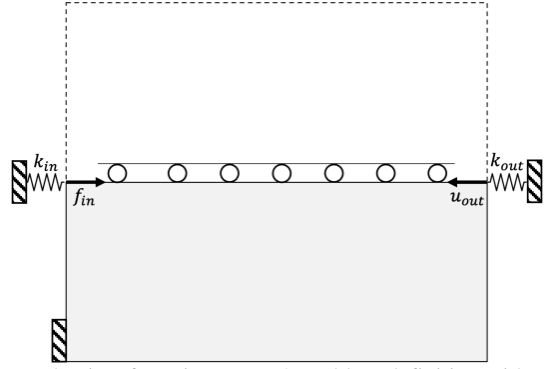

Figure 9: Compliant mechanism force inverter TO problem definition with symmetry conditions.

As illustrated in Figure 9, this benchmark TO problem [31] is defined by applying a force in the x-direction at the center of the left boundary ($f_{in}$) with the optimization objective of maximizing the negative x-displacement at the center of the right boundary ($u_{out}$). This structure is difficult to interpret because the model must accurately replicate the hinges of the mechanism to preserve its compliant force inversion structural response.

## 2.2 The YOLOv8-TO Model

YOLOv8, an anchor-free CNN computer vision model [32], consists of a CSP-Darknet53 backbone and neck architecture [33]. As shown in Figure 10, split heads generate instance segmentation outputs, including bounding box prediction, object classification, and pixel classification. A custom YOLOv8 implementation was developed to predict design variables for each detected structural component. The new regression head uses convolutional layers with ReLU activation, followed by a final convolutional layer to produce a 5D vector of normalized design variables, which are then scaled using minimum and maximum bounds (Appendix B). The regression predictions are made at multiple scales of the feature maps, allowing the model to utilize different feature scales. Nano, medium, and extra-large (x-large) models with 4.85M, 29.18M, and 73.38M parameters, respectively, were trained to assess the impact of model size on performance. Training continued until the fitness metric (Equation 1), a weighted sum of the mean Average Precision (mAP) at different Intersection over Union (IoU) thresholds, did not improve for 100 epochs. The $mAP_{@0.5-0.95}$ represents the average mAP over IoU thresholds from 0.5 to 0.95 with a step size of 0.05, comprehensively assessing the model's performance across various levels of overlap between the predicted and ground truth bounding boxes or masks. Data augmentation from the Albumentations library [34] was used during training, with parameters and hyperparameters listed in Appendix B.

$$fitness = \left(0.1\, mAP_{@0.5\,(Bbox)} + 0.9\, mAP_{@0.5-0.95\,(Bbox)}\right) + \left(0.1\, mAP_{@0.5\,(Mask)} + 0.9\, mAP_{@0.5-0.95\,(Mask)}\right) \qquad (1)$$

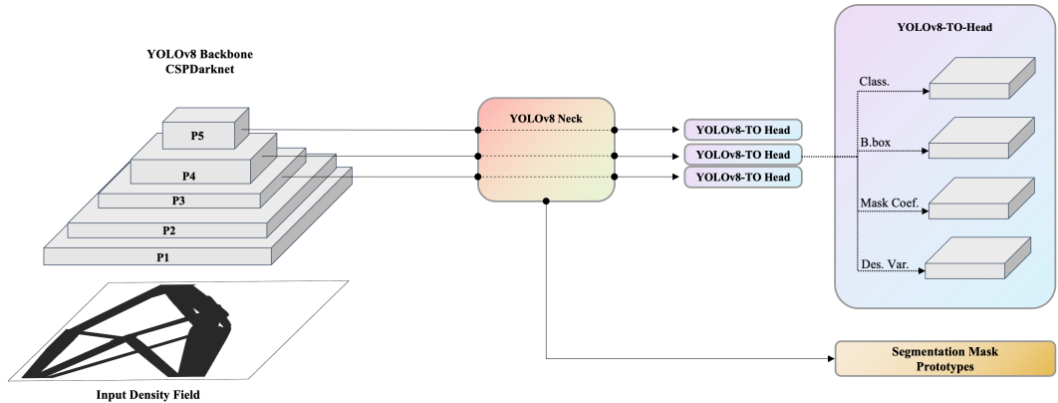

Figure 10: Architecture of the YOLOv8-TO model. The input density field is processed by a CNN CSPDarknet feature extractor (left). The neck architecture (middle) passes three different layers from the backbone to separate YOLOv8-TO heads and a segmentation mask prototype network. Each YOLOv8-TO head (right) generates predictions for classification, bounding box, mask coefficients, and design variables. The segmentation mask prototype network is a fully convolutional network that generates prototype segmentation masks.

During training with the original MMC formulation from [23], very large gradient values occurred due to the use of non-smooth projection methods. The implementation of the level set function and its smoothed Heaviside projection resulted in abrupt changes in the gradient flow, causing the model training to become unstable. To address this, a simplified MMC approach was adopted, utilizing endpoint coordinates and a single thickness value. This method circumvents the issue of angle prediction with a sigmoid function, which can misrepresent similar angles. The two approaches are shown in Figure 11 and Figure 12. A normalized sigmoid projection was employed to convert the level set function of each component into a binary density field, as detailed in Equations 2 to 9. Figure 13 illustrates the sigmoid projection scheme for a single component.

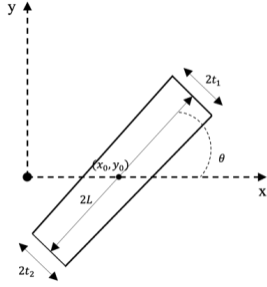
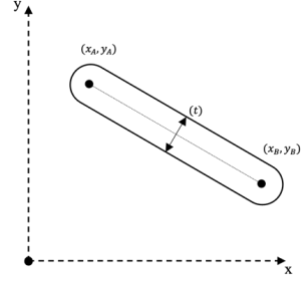

Figure 11: MMC Component formulation used in [23].

Figure 12: Endpoint MMC formulation used in this paper.

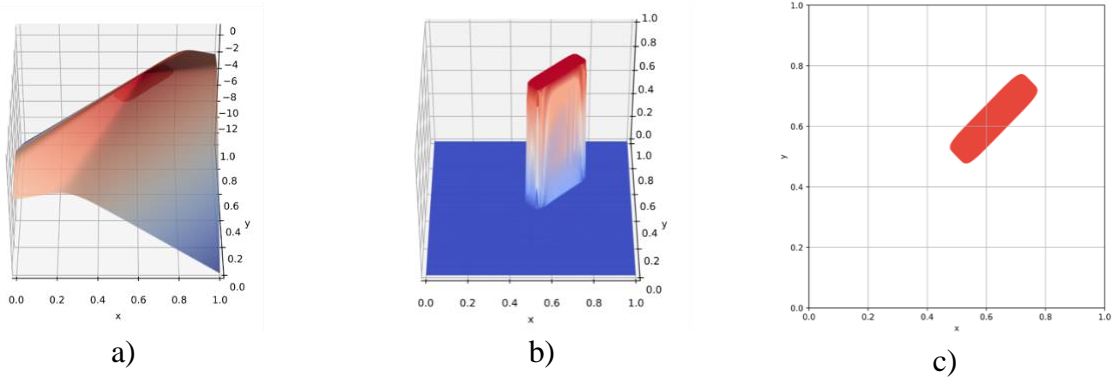

a)

b)

c)

Figure 13: Illustration of the MMC component. a) The contour surface of the level set function. Everything above the '0th' level is highlighted in red. b) The corresponding sigmoid projection. c) 2D view of the sigmoid projection corresponding to the density field of the single MMC component.

$$x_0 = \frac{A_x + B_x}{2}, y_0 = \frac{A_y + B_y}{2} \tag{2}$$

$$L = \sqrt{(B_x - A_x)^2 + (B_y - A_y)^2} \tag{3}$$

$$\theta = arctan2\,(B_y - A_y, B_x - A_x) \tag{4}$$

$$x_1 = \cos(\theta)\,(X - x_0) + \sin(\theta)(Y - y_0) \tag{5}$$

$$y_1 = -\sin(\theta)\,(X - x_0) + \cos(\theta)(Y - y_0) \tag{6}$$

$$\Phi(x_1, y_1) = 1 - \left(\left(\frac{x_1}{\frac{L}{2} + \epsilon}\right)^6 + \left(\frac{y_1}{t + \epsilon}\right)^6 + \epsilon\right)^{1/6} \tag{7}$$

$$z = \frac{\Phi(x_1, y_1) - \alpha}{\beta} \tag{8}$$

$$\rho(x_1, y_1) = \frac{1}{1 + e^{-z}} \tag{9}$$

$x_0, y_0, L$ and $\theta$ are the coordinates of the centroids, the length, and the orientation of the MMC component, defined by the endpoint coordinates $A_x, A_y, B_x$ and $B_y$. These geometric variables define the shape of the hyperelliptic function $\Phi(x_1, y_1)$. $\epsilon$, $\alpha$ and $\beta$ are normalizing constants and the final density field is obtained via the sigmoid projection $\rho(x_1, y_1)$.

### 2.2.1 Pixel-based bounded reconstruction loss

The presence of overlapping components in the MMC-generated training dataset poses a challenge for the model's training process. In the MMC method, overlapping components can combine into a larger structural component with a single visible boundary. Using a classic regression loss such as the mean-squared error (MSE) on the design variables would force the model to predict the design variables of these overlapping components, which is an undesirable behavior. To mitigate this the new design variables regression head is trained using a dice loss bounded within the predicted bounding box. The pseudocode for the loss function is presented in Figure 14. One important advantage of using such reconstruction loss is that it is inherently self-supervised, requiring no ground-truth design variables for training.

**Pixel-based bounded reconstruction loss:**
 **for** img in batch:
  $\hat{y} = YOLOv8\,(img)$   (Normalized design variables prediction in the local bounding box reference system)
  $\hat{y} \to \tilde{y}$   (Scale the design variables using the bounding box coordinates)
  $\Phi = MMC(\tilde{y})$   (Generating MMC level set functions for each component)
  $\rho = sigmoid(\Phi)$   (Sigmoid projection)
  $img \to mask_{cropped}$   (Crop corresponding pixel mask using the predicted bounding box)
  $intersection = \sum (\rho \cdot mask_{cropped})$
  $dice = \dfrac{2 \cdot intersection}{\sum \rho(\Phi) + \sum mask_{cropped}}$
 **return** $(1 - dice)/batchsize$

Figure 14: Pixel-based bounded reconstruction loss algorithm.

### 2.2.2 Segmentation mask non-max suppression

In the original YOLOv8 implementation, non-max suppression (NMS) is performed using bounding boxes during inference. However, this poses a challenge for detecting structural patterns, as symmetrical structures might share identical bounding boxes yet consist of two separate elements. An example of this is shown in Figure 15 with an equal-length orthogonally intersecting pair of components, a configuration that is commonly found in topology-optimized structures due to its high stiffness. To address this, a NMS algorithm based on the dice coefficient of the predicted mask of the model was implemented. Each mask pair is compared and overlapping pairs with a dice coefficient above a certain threshold are marked as too similar, with only the mask with the highest confidence score kept in the final prediction. This operation is, however, computationally expensive, as demonstrated by the example in Figure 15, where the bounding box NMS requires 2 milliseconds, while the dice coefficient NMS takes 82 milliseconds, highlighting the trade-off between improved accuracy and computational cost.

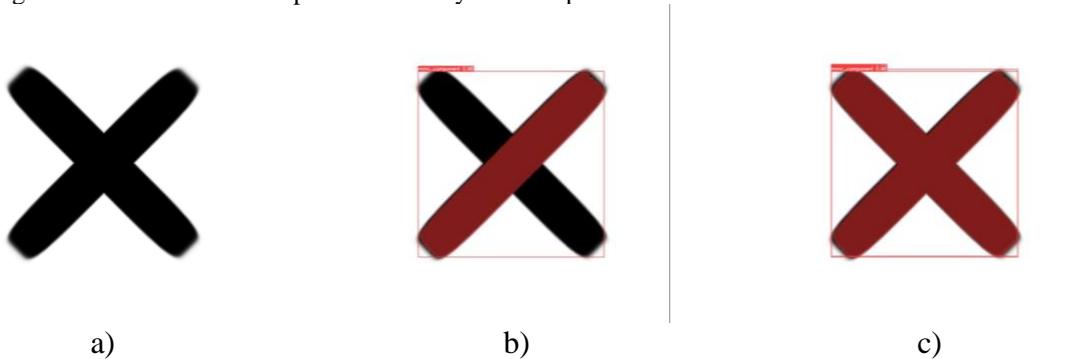

a)      b)      c)

Figure 15: a) Orthogonally intersecting pair of equal length MMC component. b) Model detection results with NMS on the bounding box. c) Model detection results with NMS based on the Dice coefficient of masks.

## 2.3 Comparing YOLOv8-TO to skeletonization

Results from the YOLOv8 approach will be compared to the skeletonization approach proposed in [16]. A thickness estimation algorithm was developed to obtain a similar parametric reconstruction as the MMC approach of YOLOv8-TO. For each detected skeleton branch, a constant thickness is estimated with the proposed algorithm described in Figure 16.

| **Skeleton thickness estimation:** | |
|---|---|
| $\text{dist} = \text{dist.transform}(\text{input image})$ | (Distance transform matrix of the input binary image using Euclidean distance and kernel size of 5) |
| **for** branch in skeleton: | (Loop through each branch) |
|     **for** point in numPoints: | (Determine a set of equally spaced points along the branch) |
|         $\text{diameter}[\text{point}] = 2 \cdot \max(\text{kernel}_{3x3}(\text{dist}[\text{point}]))$ | (Calculate the maximum distance of a 3x3 kernel centered around the point) |
|     $\text{thickness}[\text{branch}] = \text{mean}(\text{diameter})$ | (Assign the branch thickness as the mean of the calculated diameters) |
| $\text{thickness}_{5\%} = $ 5th percentile of thicknesses | (Determine the 5$^{\text{th}}$ percentile thickness as minimum threshold) |
| $\text{thickness} = \max(\text{thickness}, \text{thickness}_{5\%})$ | (Apply the minimum thickness) |
|  **return** $thickness$ | |

Figure 16: Skeleton thickness estimation algorithm.

## 2.4 Disconnected load path in predicted structures

A common issue with image-based prediction models is their tendency to generate structures with disconnected load paths. Image similarity loss functions alone do not ensure that the generated structure maintains contact with the support and loaded regions [25]. To address this problem and enable a fair comparison of structural performance between the original image and the predicted structures from YOLOv8-TO and skeletonization algorithms, a simple connection algorithm was developed.

The connection algorithm treats the support and load conditions independently, verifying that at least one pixel of material is in contact with each boundary condition to ensure the structure's connectivity. In cases where disconnection is detected, the algorithm identifies the nearest skeleton or MMC node to the initial pixel adjacent to the boundary condition. The nearest node is determined based on its row and column order in the image. The algorithm then repositions the node to establish a connection with the boundary condition.

It is important to note that the algorithm does not perform any post-processing to check for internal disconnections within the structure. Addressing internal disconnections would require the development of a more complex heuristic to identify and correct such issues. The current implementation focuses solely on ensuring connectivity at the support and load boundaries, which is sufficient for comparing the structural performance of the original and predicted structures.

## 3. Results

YOLOv8-TO and the skeletonization method were evaluated based on their ability to reconstruct the original image accurately. Three metrics were used to assess the reconstruction quality:
- Dice coefficient, which measures the image similarity between the reconstructed and input images.
- Relative volume difference, which quantifies the similarity in volume between the reconstructed and input images.
- Relative compliance difference, which assesses the similarity in structural response between the reconstructed and input images.

The Dice coefficient and volume similarity were calculated using mean values across the test datasets. In contrast, the median compliance was used due to the presence of a few internally disconnected structures that can cause a highly skewed structural response distribution. Furthermore, when the reconstructed structures' volumes are comparable between the two methods, analyzing the product of compliance and volume offers valuable insights into their performance from a structural optimization perspective. However, it is important to note that this additional metric is only valid for comparing structures with similar volumes. As a result, this metric will only be used when analyzing individual results within this section, where the volume similarity between the input and reconstructed structures is ensured.

The aggregated results, presented in Table 2, were obtained using the x-large YOLOv8-TO model trained on a combined dataset of MMC-optimized structures and random assemblies. To determine the Intersection over Union (IoU) and confidence thresholds for the NMS post-processing step, a subset of 10 randomly selected samples was used. Model predictions were generated for various IoU and confidence threshold combinations, and the combination that yielded the lowest mean dice loss across all 10 samples was selected. This optimal combination was then applied to run inference on the entire corresponding test dataset. The specific confidence and IoU combinations used for each dataset are indicated under the dataset name in Table 2. On the three MMC, SIMP & SIMP$_{5\%}$ test datasets, YOLOv8-TO yielded an average Dice coefficient improvement of 13.84%, with a maximum improvement of 20.78%.

Table 2: Aggregated results of YOLOv8-TO vs skeletonization on all test datasets.

| Method | Metrics | Evaluation Datasets | | | | | | |
|---|---|---|---|---|---|---|---|---|
| | | MMC @ (0.01, 0.5) | SIMP @ (0.01, 0.45) | SIMP$_{5\%}$ @ (0.01, 0.45) | OOD Femur total vol. | OOD Femur local vol. | OOD Level set 20 iter. | OOD Comp. Mech. |
| YOLOv8-TO | Mean Dice Coefficient [1.] | **0.94** | **0.94** | **0.90** | **0.94** | 0.89 | **0.94** | **0.95** |
| | Mean Volume Δ (%) [2.] | -6.18 | +6.83 | +11.06 | +5.61 | +5.44 | **+1.04** | -5.64 |
| | Median Compliance Δ (%) [3.] | +24.91 | +12.99 | +31.12 | **+0.98** | **-6.09** | -22.76 | **-4.36**[*] |
| Skeletonization | Mean Dice Coefficient [1.] | 0.83 | 0.88 | 0.74 | 0.85 | **0.92** | 0.88 | 0.70 |
| | Mean Volume Δ (%) [2.] | -15.67 | **-2.70** | **-4.12** | **0.00** | -4.24 | +6.12 | +3.37 |
| | Median Compliance Δ (%) [3.] | +30.41 | +18.99 | +100.89 | +4.45 | +12.07 | +11.08 | -99.50[*] |

[*] The compliant mechanism force inverter's structural performance is evaluated based on the x-displacement at the right corner.
[1.] Higher is better.
[2.] Lower is better.
[3.] Closer to 0 is better.

The training curves for the x-large YOLOv8-TO model trained on three different datasets are presented in Figure 17a. The datasets include optimized MMC structures only, random assemblies only, and a combination of both. Validation was performed on a separate subset of MMC-optimized samples for each training run. The model trained solely on random samples achieved a final bounding box mean average precision at IoU thresholds in the range of 50% to 95% ($mAP_{@0.5-0.95}$) of 0.82 and a validation dice loss of 0.11. The bounding box mAP was chosen as the preferred metric over the mask mAP due to its direct involvement in the design variables prediction. The model trained on MMC-optimized structures only attained a $mAP_{@0.5-0.95}$ of 0.921 and a validation dice loss of 0.0876. The combination of random and MMC-optimal images resulted in a slightly better performance, with a $mAP_{@0.5-0.95}$ of 0.924 and a validation dice loss of 0.0866.

Figure 17b shows the training curves for three different YOLOv8 model sizes (nano, medium, and extra-large) trained on the combined dataset. The extra-large model achieved the highest $mAP_{@0.5-0.95}$ of 0.924 for the ground truth bounding box and the smallest validation dice loss of 0.086. This indicates that increasing the model capacity allows for better learning of the complex structural patterns present in the topology-optimized designs.

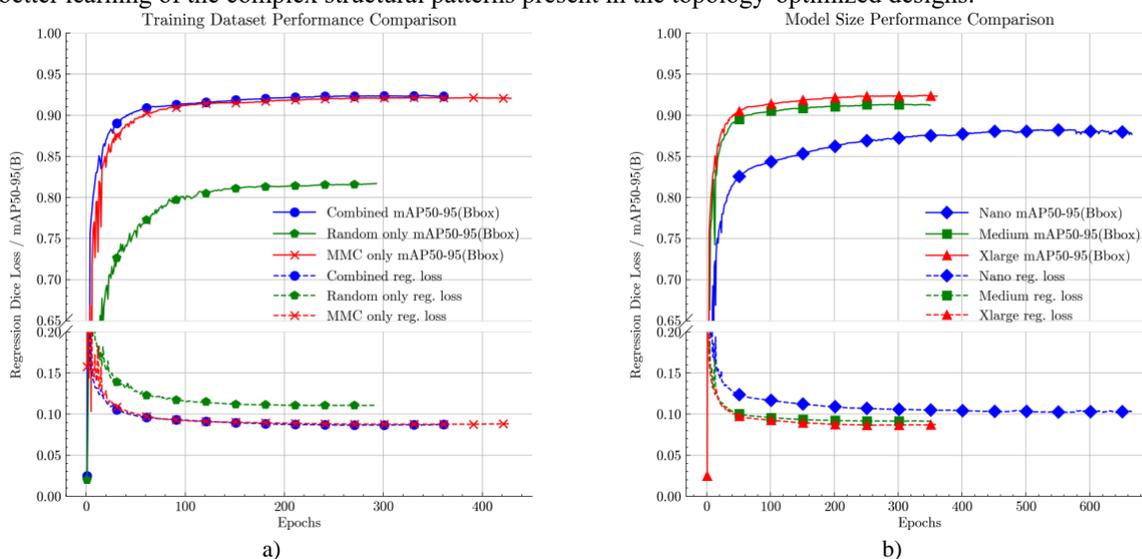

Figure 17: Comparative training curves for bounding box mean Average Precision (mAP) with a range of 50-95 and regression head loss across a) datasets incorporating MMC, random, and a combination of both, and b), various model scales.

Figure 18, Figure 19 & Figure 20 present sample results from the MMC, SIMP, and SIMP$_{5\%}$ test datasets, respectively. YOLOv8-TO, despite being trained only on MMC-optimized data, demonstrates its effectiveness in handling topologies obtained via the MMC algorithm, but also on topologies obtained with SIMP. It outperforms the skeletonization method even on truss-like structures in the SIMP$_{5\%}$ dataset.

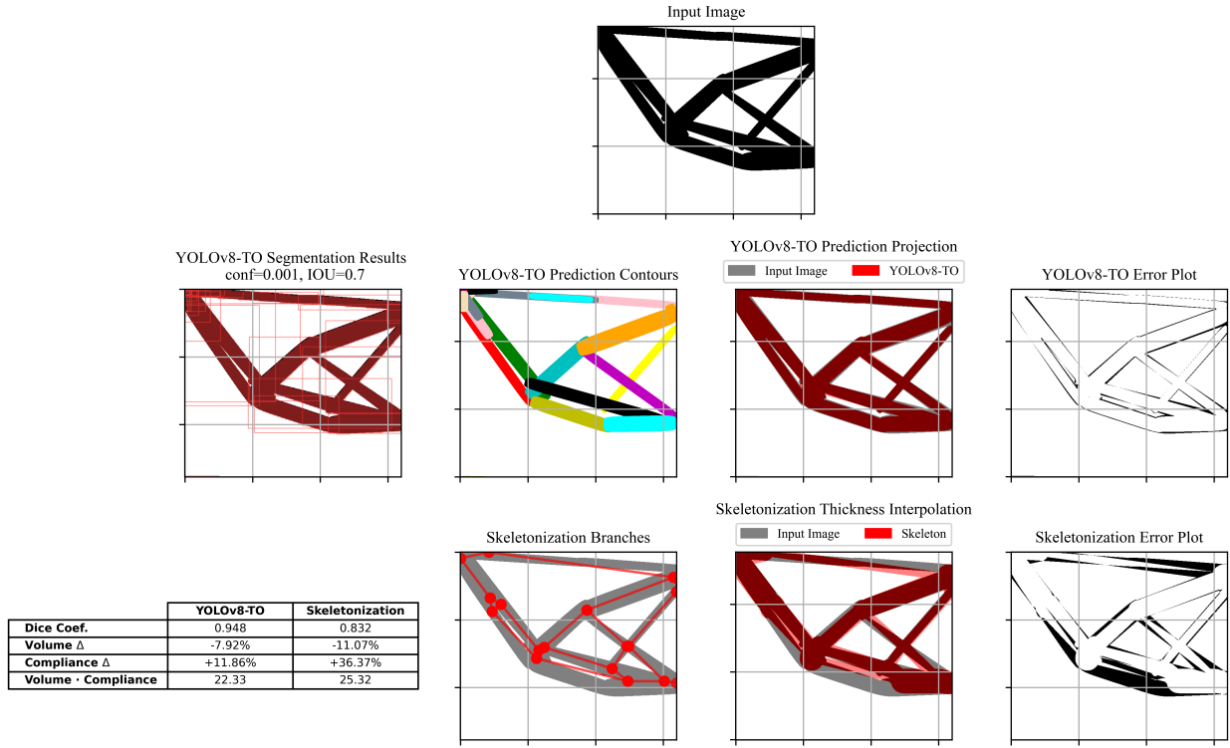

Figure 18: Results sample from the MMC test dataset.

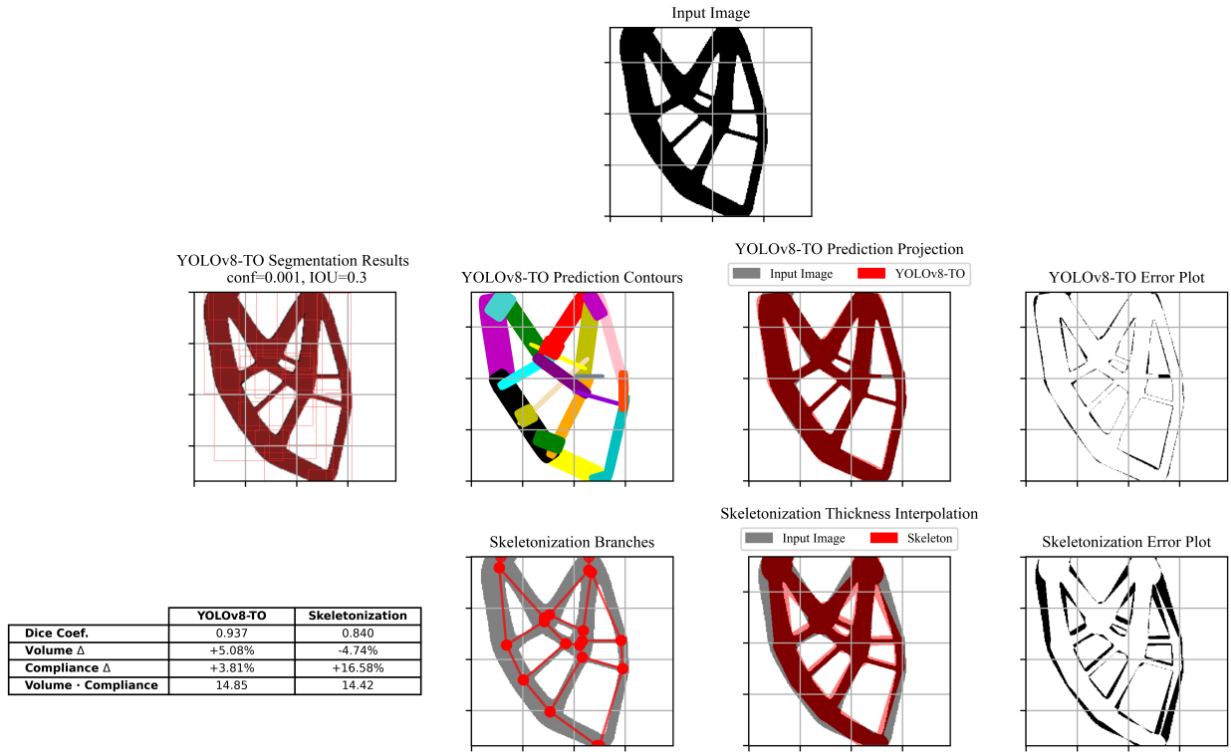

Figure 19: Results sample from the SIMP dataset.

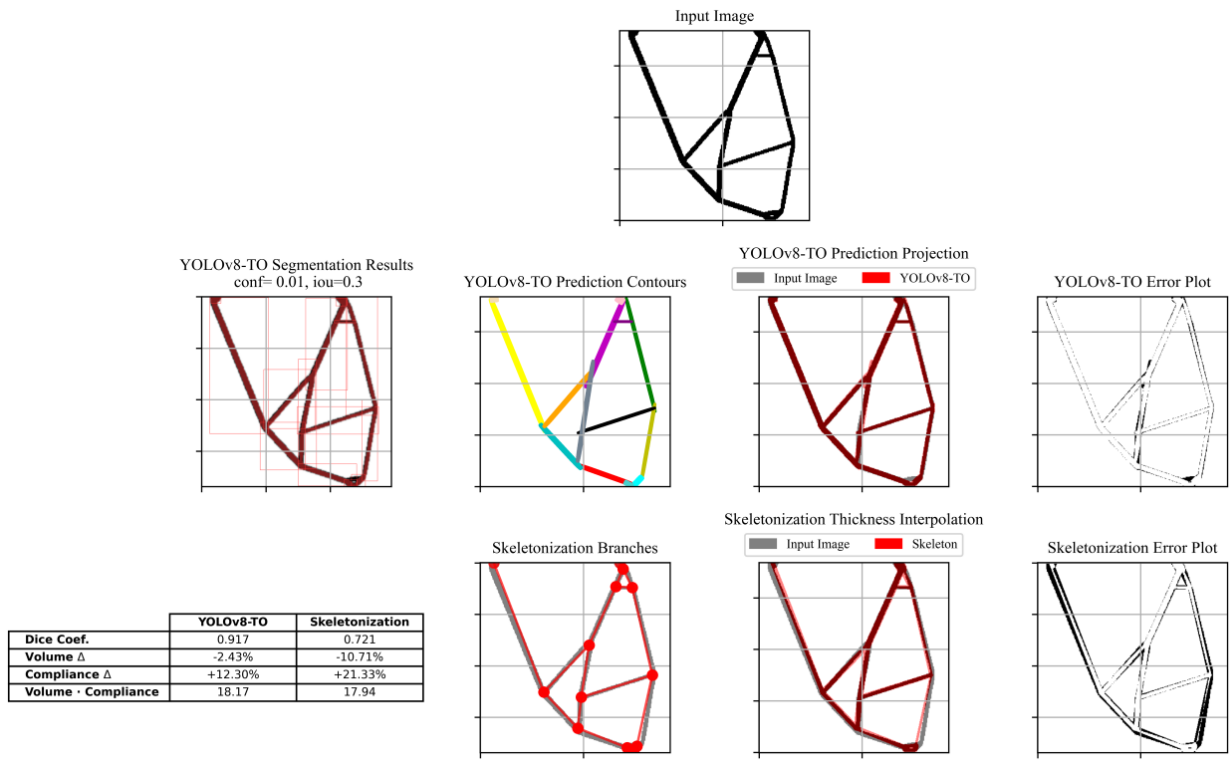

Figure 20: Results sample from the SIMP$_{5\%}$ dataset.

Figures 21 through 24 showcase the results of the OOD test samples. YOLOv8-TO achieves better image similarity compared to the skeletonization method on all OOD samples, except for the 2D femur with local volume constraint, where the skeletonization method yields a structure with a slightly higher dice coefficient.

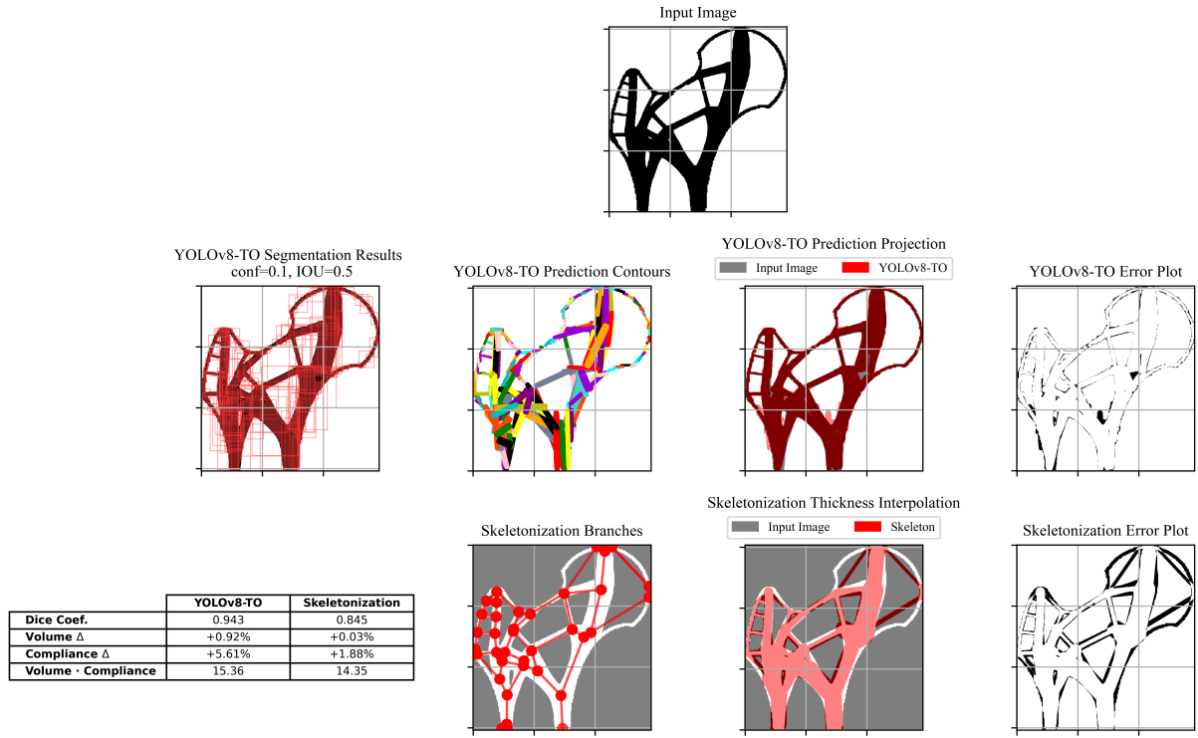

Figure 21: Results for the 2D femur case with total volume constraint.

The 2D femur example with total volume constraint, illustrated in Figure 21, demonstrates YOLOv8-TO's ability to accurately reconstruct curved structural members by assembling multiple straight MMC components, as evident in the circular top right femur head. Despite YOLOv8-TO producing a more visually similar reconstruction (Dice coefficient: 0.94 for YOLOv8-TO vs 0.85 for skeletonization), the skeletonization method outperforms YOLOv8-TO in terms of compliance (relative difference: +1.88% for skeletonization vs +5.61% for YOLOv8-TO) and volume similarity (relative difference: +0.03% for skeletonization vs +0.92% for YOLOv8-TO).

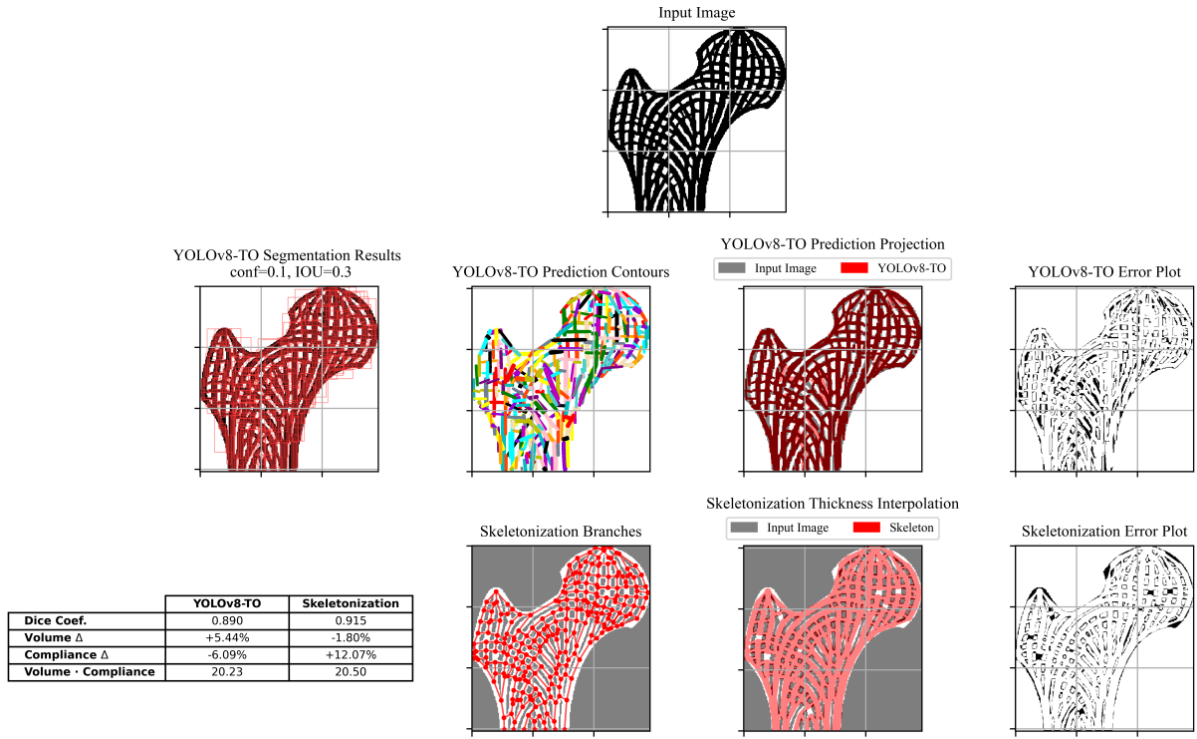

Figure 22: Results for the 2D femur case with local volume constraint.

The 2D femur with local volume constraint, shown in Figure 22, is the OOD sample where the skeletonization method surpasses the YOLOv8-TO method on image similarity (Dice coefficient: 0.92 for skeletonization vs 0.89 for YOLOv8-TO). The accurate detection of skeleton nodes, facilitated by the presence of short structural members and numerous junctions, enables the skeletonization approach to fit curves precisely. However, when considering the structural efficiency, the YOLOv8-TO reconstruction still exhibits a lower product of volume and compliance (20.23 for YOLOv8-TO vs 20.50 for skeletonization), suggesting a more optimal utilization of the reconstructed material.

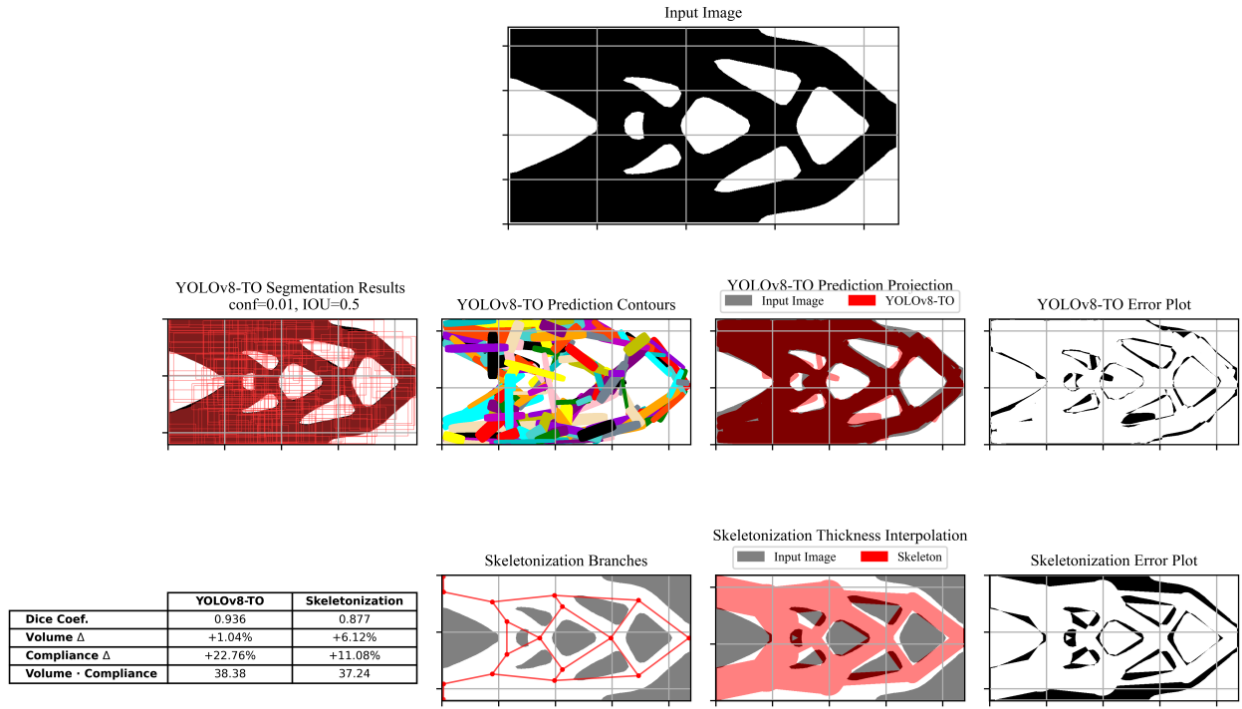

Figure 23: Results for the level set cantilever beam after 20 iterations.

The results of the partially optimized level set cantilever beam in Figure 23 reveal that the YOLOv8-TO model struggles to reconstruct structural members thicker than those encountered during its training phase. This limitation also arises from the use of an upper bound on the predicted components' thicknesses, resulting in an unnecessarily large number of components being predicted to achieve a reconstruction with a high Dice coefficient (0.94 for YOLOv8-TO vs 0.88 for skeletonization). Consequently, the YOLOv8-TO reconstruction exhibits a higher relative difference in compliance (+22.76% for YOLOv8-TO vs +11.08% for skeletonization) and volume (+1.04% for YOLOv8-TO vs +6.12% for skeletonization) compared to the skeletonization method.

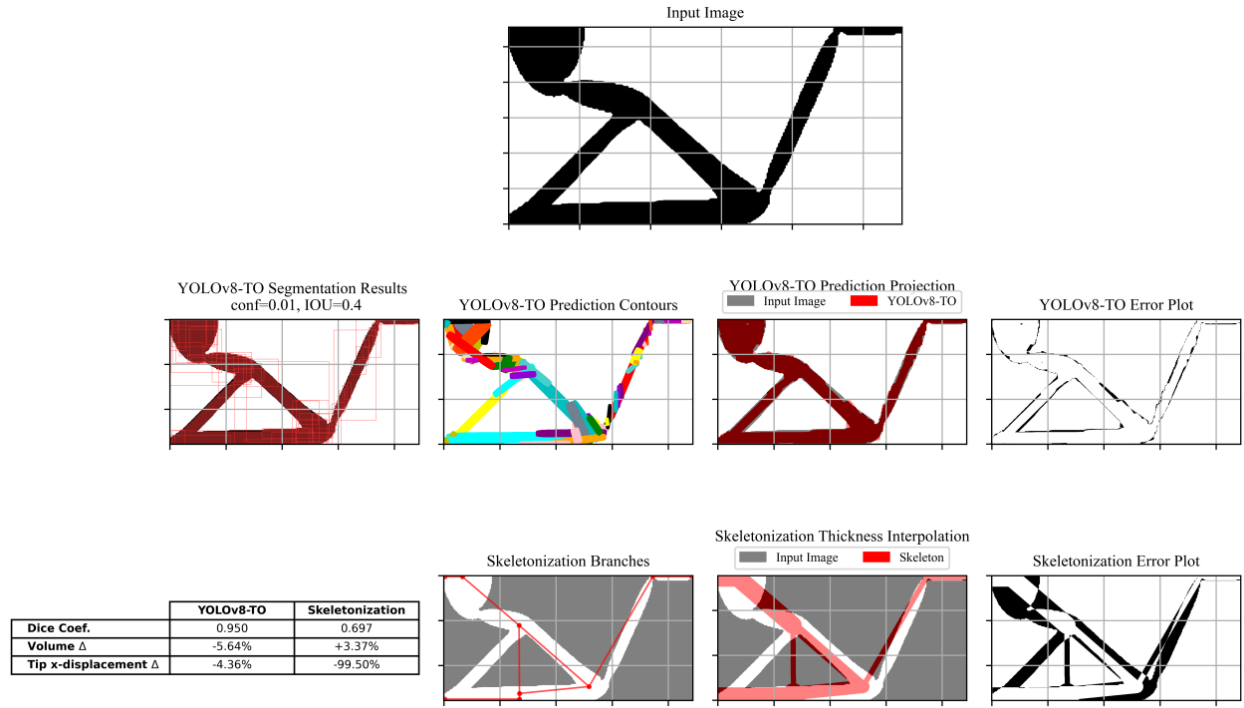

Figure 24: Results for the compliant mechanism force inverter.

The compliant mechanism results in Figure 24 demonstrate YOLOv8-TO's ability to accurately reconstruct crucial structural features, such as the mechanism's hinges, enabling the reconstructed structure to exhibit a tip displacement similar to the original design (relative difference: -4.36% for YOLOv8-TO vs -99.50% for skeletonization). In contrast, the skeletonization method fails to detect skeleton nodes at the hinge locations, resulting in straight bars where a thinning of the structure would normally be required to allow bending. Consequently, the skeletonization approach produces a stiff reconstructed structure that does not capture the essential characteristics of the input design. Additionally, the YOLOv8-TO reconstruction achieves a higher Dice coefficient (0.95 for YOLOv8-TO vs 0.70 for skeletonization), further highlighting its superior performance in reconstructing the compliant mechanism.

## 4. Discussion

In comparative analyses, YOLOv8-TO exhibits a marked superior performance over the skeletonization method, with an average Dice coefficient improvement of 13.84% in generating reverse-engineered designs on the three large test datasets. This notable advancement provides a significant increase in visual similarity to the original structures, with maximum observed enhancements reaching as high as 20.78%.

### 4.1. Improvement in precision

It was observed that the creation of skeletons typically encounters issues with the misplacement of nodes during the extraction process, primarily due to the reliance on geodesic distances for the skeleton node detection. Such an approach tends to place nodes at the centers of structural "centroids" rather than at the center of the actual junctions, leading to inaccuracies. An example from the MMC test dataset, illustrated in Figure 18, demonstrates a significant misplacement of the lower skeleton members by the skeletonization algorithm, resulting in a notable deviation between the reconstructed and the original structure. In contrast, the YOLOv8-TO reconstruction is remarkably precise, with only minor discrepancies in the thickness of the members that marginally affect structural compliance. The results from Table 2 validate the effectiveness of the thickness estimation algorithm described in Figure 16, as the skeletonization method produced structures with more similar volumes due to the use of mean diameter calculations to predict the final diameter of each skeleton branch. When considering only the compliance-volume product and disregarding the visual similarity to the original structure, the skeletonization method often produces structures with lower or similar values compared to YOLOv8-TO on the individual samples of the results section, suggesting its effectiveness in generating more structurally efficient designs.

### 4.2 Small impact from using random structures

The combination of randomly generated MMC assemblies, such as those illustrated in Figure 5, with optimal structures resulted in a 1.14% decrease in validation dice loss and a 0.33% increase in $mAP_{@0.5-0.95}$. Although using only random structures severely underperforms compared to using MMC-only or combined datasets, the marginal performance improvement observed with the combined dataset suggests that the inclusion of random samples can be beneficial. While the higher training cost associated with processing a larger number of training samples at each epoch for the combined dataset is a consideration, it can be argued that the one-time training cost is less important than the potential long-term benefits of a better-performing model. The improved performance, even if small, could lead to more accurate and efficient results when the model is deployed and used extensively in real-world applications. Future work will explore the potential of using more structured noise to improve performance while minimizing the impact on training costs.

### 4.3 Scaling yields a significant performance boost

Scaling the model size upwards led to notable improvements in performance metrics and training efficiency. Transitioning from the nano-sized model to the medium-sized one yielded a 3.50% enhancement in the validation bounding box $mAP_{@0.5-0.95}$, coupled with a 10.41% reduction in reconstruction dice loss. As an additional scaling benefit, the larger models converged much faster, requiring almost half the number of epochs (350 vs 650) to reach the same level of performance as the nano model. Furthermore, upgrading from a medium-sized model to an extra-large variant resulted in a 1.20% increase in $mAP_{@0.5-0.95}$ and a 5.29% diminution in reconstruction dice loss.

### 4.4 Faster inference speed on common workspace hardware

A comparison of inference cost was made by deploying the three tested YOLOv8-TO model sizes and the skeletonization method on a MacBook Air M2 CPU. Table 3 presents the processing times on the two 2D femur OOD test samples with total and local volume constraints, which were processed at resolutions of 736x640 and 736x672 pixels respectively. The inference time of the skeletonization method does not include the thickness estimation algorithm. The skeletonization method's inference cost exhibits a substantial increase (5.50s. vs 53.58s.) when moving from total volume constraint to local volume constraints, indicating that its processing time scales with the number of detected skeleton members. In contrast, the relatively constant inference processing time between the two images for the YOLOv8-TO models demonstrates that the forward pass through the model scales with the image resolution. The data also highlights the significant processing time required for mask-based NMS operations, emphasizing the costly nature of pixel-wise operations for each detected component. The paper's implementation uses a basic batched loop over the tensor of detected components, which lacks performance optimization. This approach results in scalability issues as the number of components increases, a problem highlighted by the increasing processing times observed across the two test samples. Nevertheless, once trained, YOLOv8-TO achieves faster inference speed on the CPU than the skeletonization method and can offer a balance between processing speed and accuracy depending on whether the bounding box or the mask-based NMS operation is selected at the post-processing step.

Table 3: Comparison of inference cost of YOLOv8-TO vs Skeletonization method.

| Test sample | Model/Method | Inference processing time (s.) | Inference + Box NMS (s.) | Inference + Mask NMS (s.) |
|---|---|---|---|---|
| 2D femur with total volume constraint | YOLOv8-TO Nano | 0.56 | 0.62 | 20.12 |
| | YOLOv8-TO Medium | 1.35 | 1.39 | 20.49 |
| | YOLOv8-TO X-large | 2.76 | 2.80 | 17.49 |
| | Skeletonization | 5.50 | - | - |
| 2D femur with local volume constraint | YOLOv8-TO Nano | 0.57 | 0.76 | 153.55 |
| | YOLOv8-TO Medium | 1.36 | 1.54 | 172.42 |
| | YOLOv8-TO X-large | 2.83 | 2.98 | 186.48 |
| | Skeletonization | 53.58 | - | - |

## 4.5 Sensitivity issues

One significant challenge with YOLOv8-TO is its strong reliance on the confidence and IoU thresholds during the NMS post-processing step, which affects the number of reconstructed structural elements. The confidence threshold determines the minimum predicted probability required for a detected component to be considered valid, while the IoU threshold is used to suppress overlapping components. Higher confidence thresholds lead to fewer detected components, as only the most confident predictions are retained, reducing false positives but potentially increasing false negatives. Similarly, higher IoU thresholds result in more aggressive suppression of overlapping components, reducing false positives but possibly eliminating true positive detections. Balancing these thresholds is crucial to minimize both false positives and false negatives in the reconstructed structure. Figure 25 illustrates the sensitivity of the interpreted structure to different confidence and IoU threshold pairs for the level set OOD example, highlighting the need to test various values when reconstructing an image to find the optimal balance. The number of interpreted components is also higher than the skeletonization output, as shown in Figure 23, which is a recurrent weakness when applying YOLOv8-TO to "thick" structures. To improve the performance of the YOLOV8-TO model, one will need to address the following three challenges:

- The first challenge is the model's reliance on predefined design variable min/max boundaries, which limits the maximum thickness of individual predicted elements. This limitation often requires the aggregation of multiple components to mimic the desired thickness, suggesting that expanding these boundaries could be a promising direction for future exploration.

- The second challenge stems from the constraints of only using straight MMC components during training, forcing the model to combine multiple straight components to reconstruct curved segments. Figure 21 exemplifies this issue, where the model assembles numerous tiny MMC components to reconstruct the curved femoral head in the upper right corner. Incorporating curved MMC components [35], which are supported by the YOLOv8 framework as a separate class of object, will be investigated in future work. Nevertheless, the model's ability to achieve this level of reconstruction using only MMC-optimized and random MMC data is remarkable.

- The third challenge arises from the presence of overlapping components in the MMC-optimized training dataset, which forces the model to predict the presence of hidden components during training. This is a consequence of the MMC method's need to "hide" or move components out of the domain to remove their contribution to the overall design. Implementing a data processing step to remove hidden MMC components or employing a feature-mapping method that handles intermediate densities [36] could prevent the model from being forced to detect redundant overlapping structures.

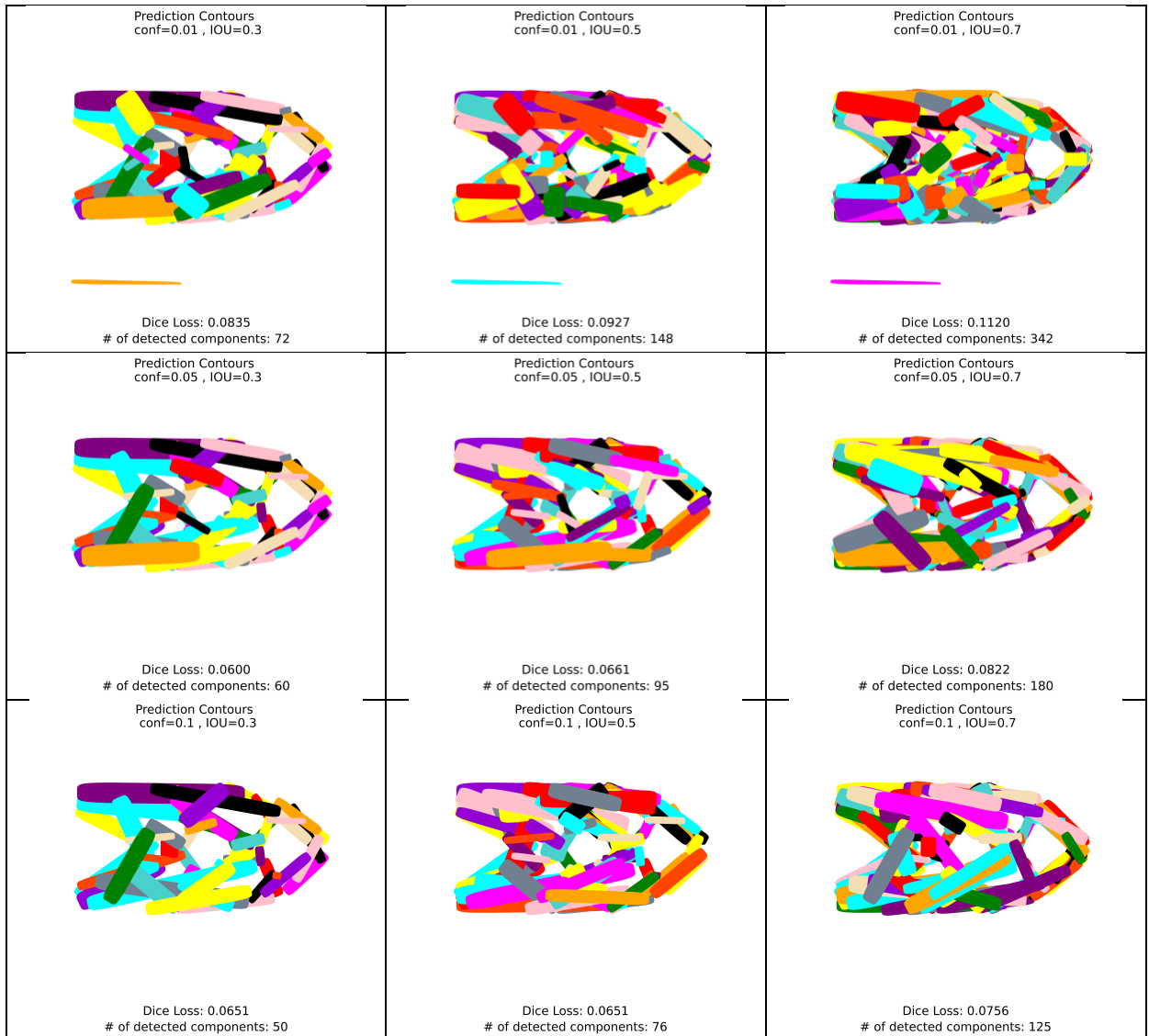
Figure 25: Effect of confidence and IoU threshold on detected components for the OOD level set example with YOLOv8-TO x-large.

## 5. Conclusion

This paper presents YOLOv8-TO, a novel approach that harnesses the power of the YOLOv8 computer vision neural network model to bridge the gap between density-based TO outputs and interpretable geometric parameters. As the first application of an instance segmentation model in the field of TO, this approach significantly enhances the post-processing stage of structural optimization, enabling a more seamless integration with subsequent design validation and exploration processes.

Our findings demonstrate that YOLOv8-TO can accurately identify and reconstruct the shapes of structural components within monochrome images of optimized structures generated from various TO methods, providing a geometrically meaningful representation that is directly usable within CAD tools. The trained model was shown to be a substantial advancement over traditional skeletonization methods, particularly in its ability to generate visually similar reverse-engineered structures. One significant advantage of this approach is that the size of the design variable regression head is independent of the input image resolution, thanks to the use of feature-mapping methods. This means that the model can predict the same number of design variables regardless of the resolution of the input image, making it a notable improvement over other machine learning approaches to TO that are often bound to coarse density resolutions. The superiority of YOLOv8-TO in handling a diverse range of structural geometries, combined with its low inference latency, underscores its potential to significantly streamline the design process in engineering applications.

The high performance and ease of use of YOLOv8-TO open the possibilities for innovative workflows in the structural design process. One such interesting application is the ability to convert hand-drawn sketches of structures into parameterized geometry by applying the model to images of such sketches. Appendix A showcases examples of doodles created using a laptop trackpad and a pen on paper, along with their corresponding reverse-engineered structures generated by the segmentation model. With YOLOv8's focus on efficiency, the trained model can be easily deployed on common workspace hardware, such as laptops or even mobile devices, enabling a seamless transition from conceptual sketches to parameterized initial designs. This ease of deployment significantly enhances the accessibility and practicality of the proposed approach, making it a valuable tool for engineers and designers in their everyday work environments.

The performance of YOLOv8-TO is constrained by the specifics of its training settings and datasets, as it currently only supports straight MMC components and has been trained exclusively on optimized and random MMC structures. Consequently, its performance may be limited when dealing with sub-structures that fall outside the dimensional bounds of the components encountered during training. Furthermore, the proposed segmentation mask non-maximum suppression (NMS) post-processing step is computationally expensive compared to the default bounding box NMS algorithm, necessitating improvements in computational efficiency before it can be applied to larger images or extended to 3D inputs.

Future research should focus on improving the generalizability of YOLOv8-TO across different structural complexities. This could involve the incorporation of more complex geometric primitives and the development of advanced data augmentation strategies to enhance OOD generalization. Enhancing the algorithm's scalability to efficiently handle 3D structures and integrating it more seamlessly with CAD and FEA tools are also critical areas for development. Moreover, exploring adaptive non-max suppression thresholds and investigating alternative loss functions could yield further improvements in the method's accuracy and reliability.

YOLOv8-TO represents a significant step forward in topology optimization post-processing. By providing an efficient and accurate means of converting density-based optimization outputs into geometrically interpretable parameters, this method holds the promise of streamlining the design process, fostering innovation, and enhancing the manufacturability of optimized structures.

# Data & Code Availability
The code and data used for training and testing the YOLOv8-TO model will be made publicly available on GitHub at https://github.com/COSIM-Lab/YOLOv8-TO. This repository contains the necessary scripts, datasets, and instructions to reproduce the results presented in this article.

# Acknowledgment
We acknowledge the support of the Natural Sciences and Engineering Research Council of Canada (NSERC) [funding reference number 569251]. This research was enabled in part by support provided by Calcul Québec (https://www.calculquebec.ca) and the Digital Research Alliance of Canada (DRAC) (alliancecan.ca).

# Appendix A – "Drawings" to geometry test

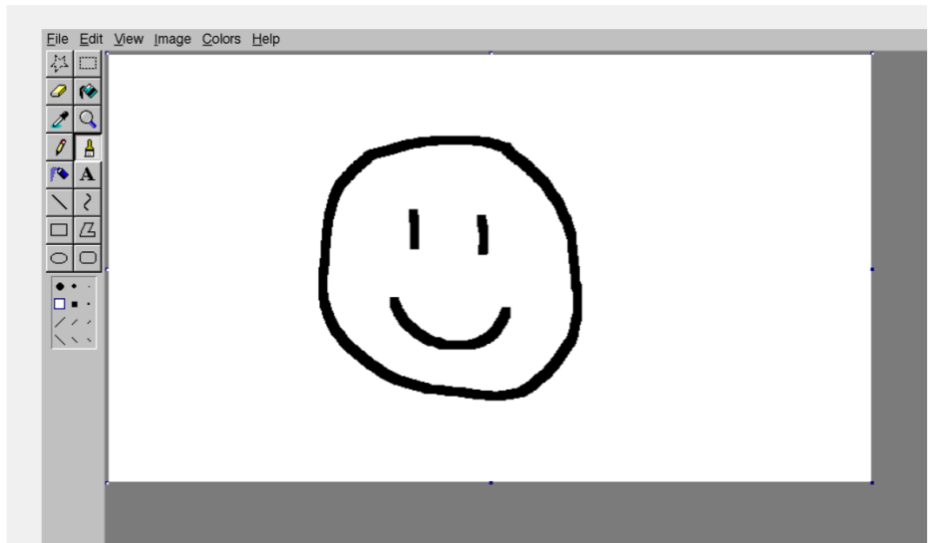

Figure 26: Smiley face computer doodle made via a laptop trackpad.

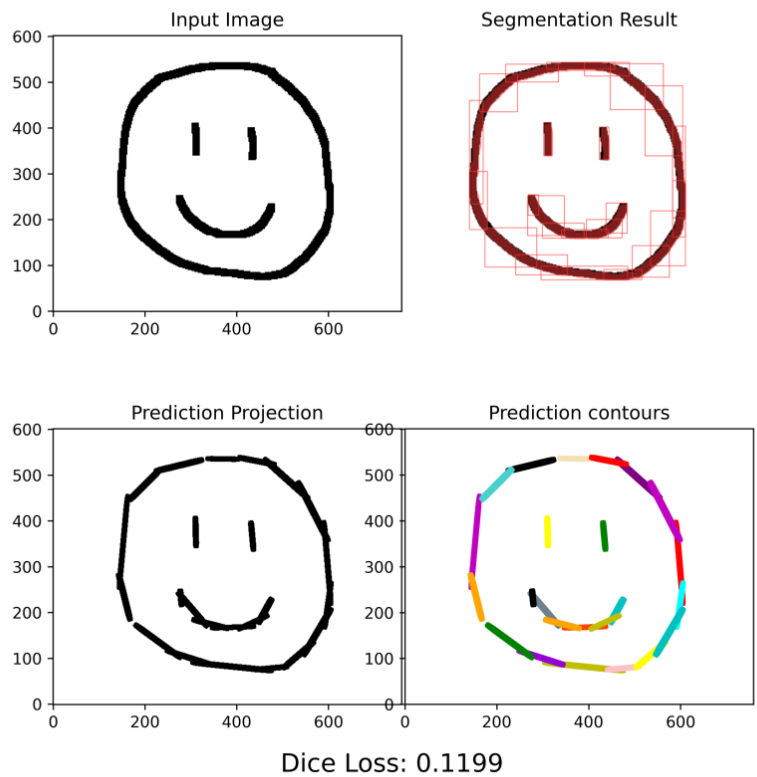

Figure 27: YOLOv8-TO reconstruction of the smiley face doodle.

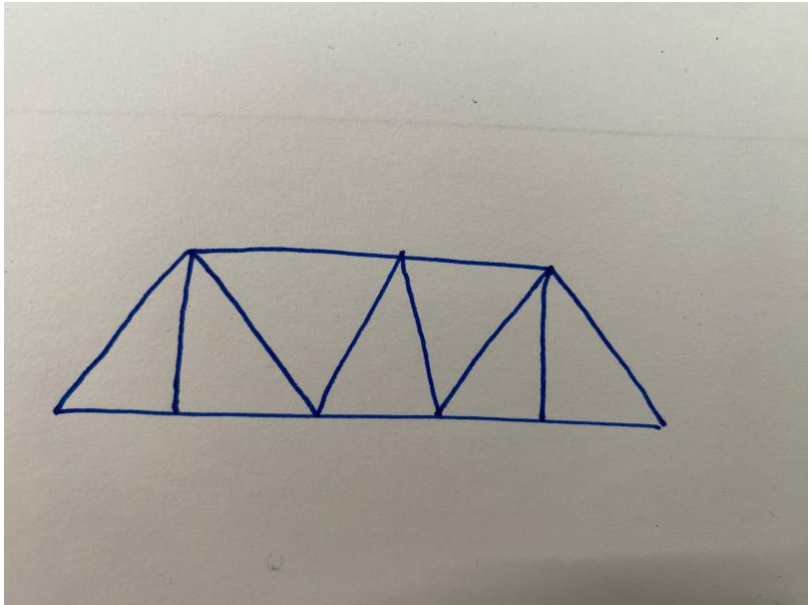

Figure 28: Cellphone image of a hand-drawn doodle of a bridge.

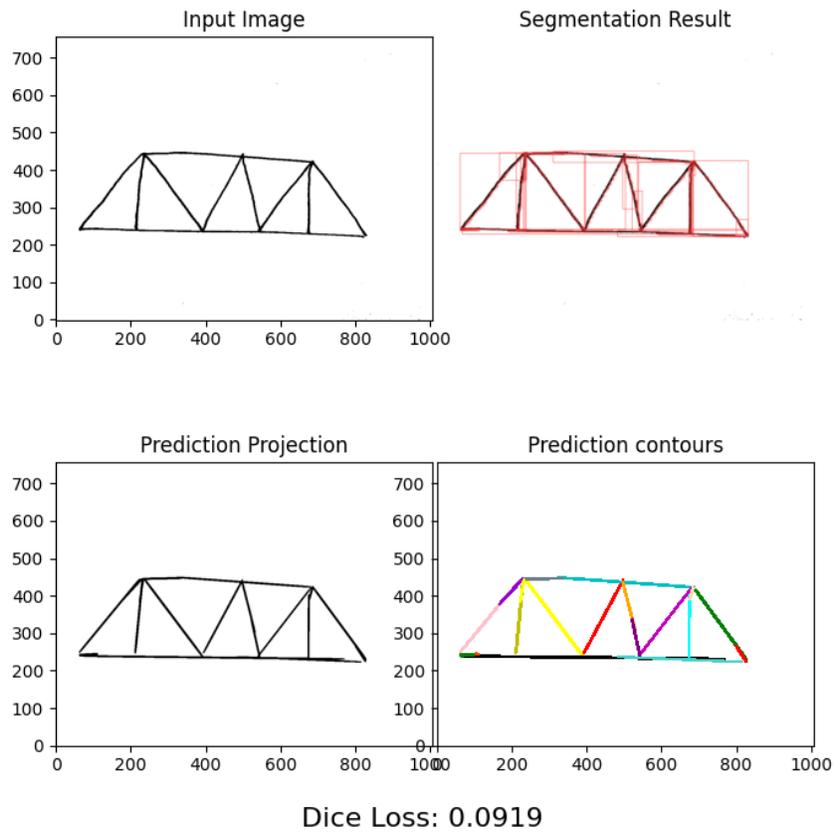

Figure 29: YOLOv8-TO reconstruction of the hand-drawn bridge image.

# Appendix B – Training Hyperparameters

The final training of all the models was done on a single compute node of the Narval compute cluster of Digital Research Alliance of Canada (DRAC), using 48 CPU cores and 4 Nvidia A100 for data parallelism distributed YOLOv8 training. Preliminary training was done on a single Nvidia 3080 GPU and 32 cores CPU on a single compute server. A cosine learning rate scheduler was used with a final learning rate factor of 1% of the initial learning rate. Table 4 details the hyperparameters used to configure the training phase of YOLOv8 that are different from their default values. Table 5 lists the minimum and maximum bounds used to normalize the design variables between 0 and 1. Table 6 lists the parameters of the data augmentation used during training.

Table 4: YOLOv8 Hyperparameters

| Hyperparameter name | Value |
|---|---|
| Per device batch size | 16 |
| Maximum number of epochs | 1000 |
| Automatic Mixed Precision (AMP) | True |
| Training image size | 640 |
| Initial learning rate | 3e-4 |
| Optimizer | AdamW |
| L2 regularization | 0.01 |
| Warmup epochs | 0 |

Table 5: design variables maximum and minimum boundaries

| Design variable | Definition | Minimum value | Maximum value |
|---|---|---|---|
| xa | X coordinate of endpoint A | Minimum x coordinate of the predicted bounding box | Maximum x coordinate of the predicted bounding box |
| ya | Y coordinate of endpoint A | Minimum y coordinate of the predicted bounding box | Maximum y coordinate of the predicted bounding box |
| xb | X coordinate of endpoint B | Minimum x coordinate of the predicted bounding box | Maximum x coordinate of the predicted bounding box |
| yb | Y coordinate of endpoint B | Minimum y coordinate of the predicted bounding box | Maximum y coordinate of the predicted bounding box |
| t | MMC thickness | 0.001 | 0.2 |

Table 6: Data augmentation parameters used during training of YOLOv8-TO.

| Augmentation type: | Probability: | Parameters: |
|---|---|---|
| Mosaic | 1.0 | 2 x 2 grid |
| Horizontal flip | 0.5 | - |
| Vertical flip | 0.5 | - |
| Image rotation | 1.0 | $U(-45°, +45°)$ |
| Image translation | 1.0 | $U(-40\%, +40\%)$ |
| Image shear | 1.0 | $U(-10°, +10°)$ |
| Image scaling | 1.0 | $U(50\%, +150\%)$ |
| Image perspective distortion | 1.0 | x/y-distortion: $U(-0.001, +0.001)$ |
| JPEG Image Compression | 0.1 | $U(75\%, 100\%)$ |
| Image blur | 0.01 | Kernel size : $U(3,7)$ |
| Median blur | 0.01 | Aperture size : $U(3,7)$ |